\documentclass{article}
\usepackage{amsmath,epsfig}
\usepackage[preprint]{spconf}
\usepackage[shortlabels]{enumitem}
\usepackage{xcolor}
\usepackage{cite}
\usepackage{amsmath,amsfonts}
\usepackage{url}

\let\OLDthebibliography\thebibliography
\renewcommand\thebibliography[1]{
  \OLDthebibliography{#1}
  \setlength{\parskip}{0pt}
  \setlength{\itemsep}{0pt plus 0.3ex}
}

\begin{document}\sloppy

\def\x{{\mathbf x}}
\def\L{{\cal L}}

\title{SE-GAN: Skeleton Enhanced GAN-based Model for Brush Handwriting Font Generation}
%
\name{Shaozu Yuan$^{1}$, Ruixue Liu$^{1}$, Meng Chen$^{1}$$^{\dagger}$\thanks{$^{\dagger}$ Corresponding author, email: chenmeng20@jd.com.}, Baoyang Chen$^{2}$, Zhijie Qiu$^{2}$, Xiaodong He$^{1}$}

\address{$^{1}$JD AI, Beijing, China $^{2}$Central Academy of Fine Arts, China}

\maketitle
\begin{abstract}
Previous works on font generation mainly focus on the standard print fonts where character's shape is stable and strokes are clearly separated. There is rare research on brush handwriting font generation, which involves holistic structure changes and complex strokes transfer. To address this issue, we propose a novel GAN-based image translation model by integrating the skeleton information. We first extract the skeleton from training images, then design an image encoder and a skeleton encoder to extract corresponding features. A self-attentive refined attention module is devised to guide the model to learn distinctive features between different domains. A skeleton discriminator is involved to first synthesize the skeleton image from the generated image with a pre-trained generator, then to judge its realness to the target one. We also contribute a large-scale brush handwriting font image dataset with six styles and 15,000 high-resolution images. Both quantitative and qualitative experimental results demonstrate the competitiveness of our proposed model.

\end{abstract}

\section{Introduction}
\label{sec:introduction}
During thousands-year history of Chinese calligraphy, many styles of writing or chirography came into being. The chirography style can be defined as the skeleton structure and stroke style. The skeleton contains the basic information of character, such as the composition and position of strokes, writing direction, etc., while the stroke style means the deformation of the skeleton, such as the thickness, shape, writing strength, etc. Intuitively, it's essential to ensure structure correct and keep style consistent when generating brush handwriting font automatically.

\begin{figure}[t]
    \begin{center}
          \includegraphics[width=.9\linewidth]{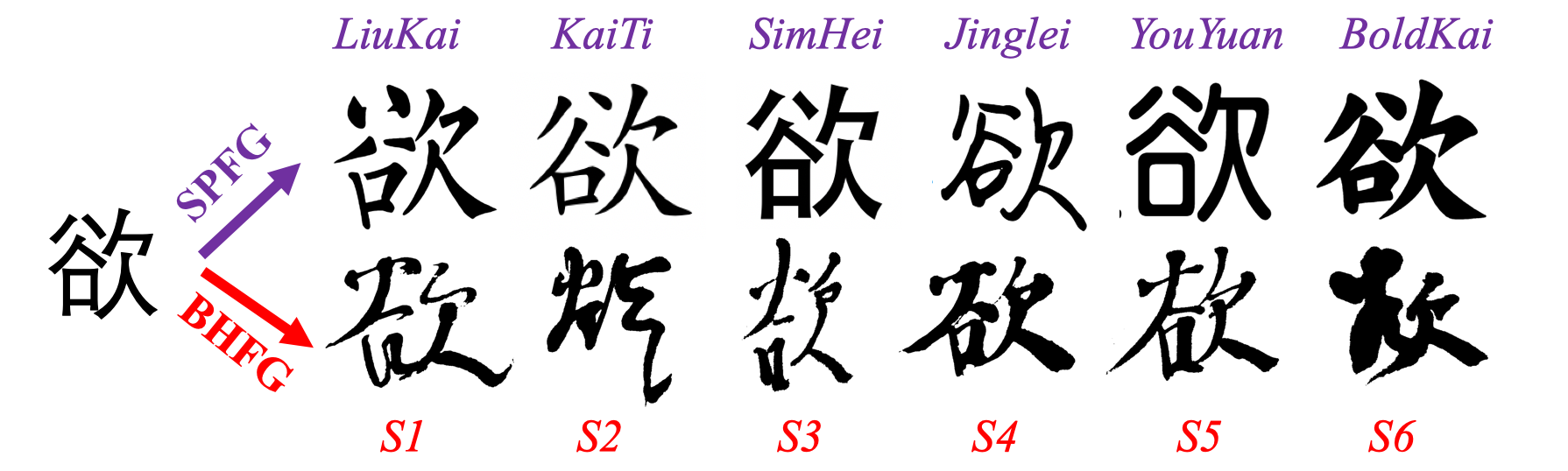}
    \end{center}
    \caption{Illustration of SPFG and BHFG.}
    \label{fig:introd}
\end{figure}

Recent works \cite{zi2zi,SAVAE2018,calligraphy_cycleGAN,gao2019artistic,wen2021stroke} formulate the Chinese font generation as an image style transfer problem, where characters in the reference style are transferred to a specific style. As it's time-consuming and labor-intensive to create a handwriting Chinese font library, most of the researches focus on Standard Print Font Generation (SPFG), however, there is rare research on Brush Handwriting Font Generation (BHFG). As shown in Fig. \ref{fig:introd}, BHFG is much more challenging than SPFG. It's observed that, even for the same character, the character images written in different styles look quite different. Especially for the characters written in cursive or semi-cursive styles, their images are heavily distorted. On one hand, the basic structure and layout of separated strokes share some similarities so that the character can be recognisable. One the other hand, there exist large geometric variations in the shape so the impressive styles can be easily distinguished. Based on these observation and analysis, we argue that the skeleton of character is critical for preserving the character content among different styles. However, most existing approaches designed for SPFG neglect the importance of skeleton. 

\begin{figure*}[ht]
	\centering 
	\includegraphics[width=0.8\linewidth]{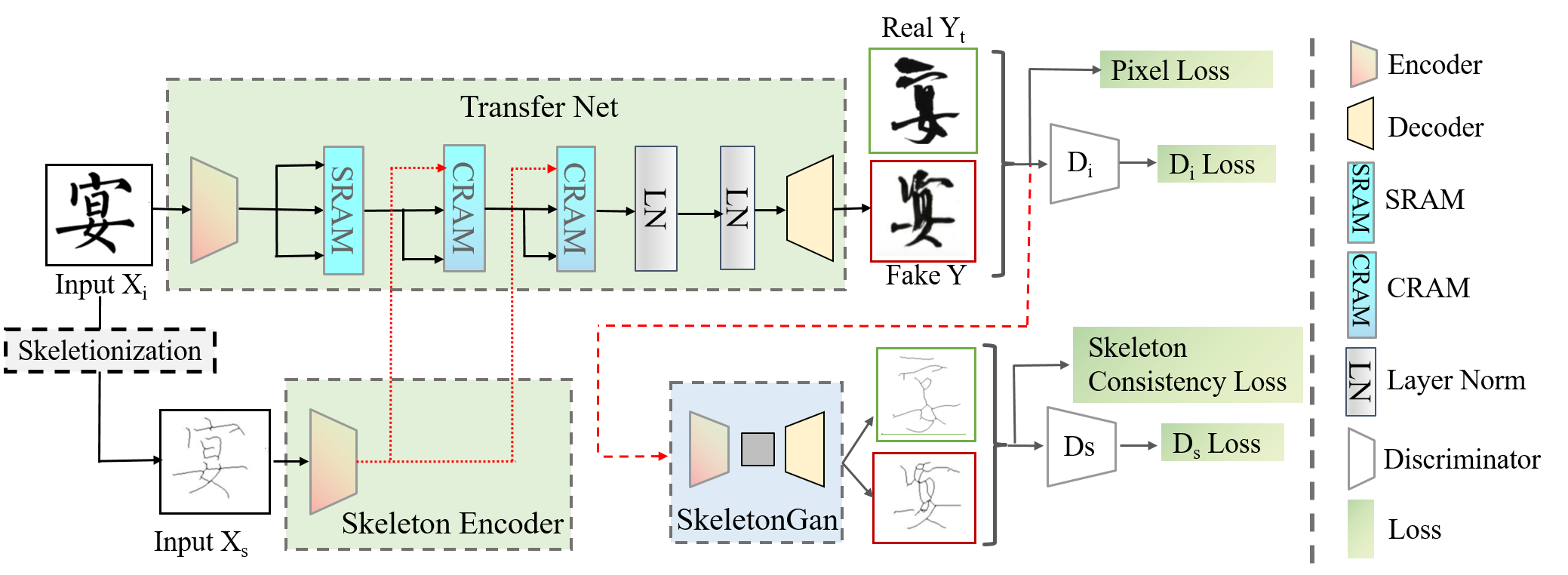} 
	\caption{The architecture of SE-GAN, including a transfer net, two discriminators, and an auxiliary skeletonGAN network.
	} 
	\label{fig:framework}   
\end{figure*}

To address above issue, we first collect a large-scale brush handwriting Chinese font dataset, which contains six different chirography styles and more than 15,000 high-resolution images. Then we propose a novel end-to-end \textbf{S}keleton \textbf{E}nhanced \textbf{G}enerative \textbf{A}dversarial \textbf{N}etwork (denoted as \textbf{SE-GAN}) for BHFG which can handle the large geometric variations between different styles. SE-GAN is a one-stage model, which means that the generator can directly output stylized images with user-specified input character. SE-GAN includes two encoders to catch the features from both source image and corresponding skeleton image. To extract and fuse the features from two sources effectively, a novel Self-attentive Refined Attention Module (SAttRAM) is devised and applied in the generator. 
To further ensure the structure preservation, we also design an extra skeleton discriminator to keep the skeleton of generated image close to the skeleton of target image. Extensive experiments were conducted on six different stylized brush handwriting font generation tasks. The experimental results show the competitiveness of our proposed model compared to the baselines.

The contributions of this paper can be summarized as follows:
1) We propose a novel end-to-end GAN-based model (SE-GAN) for BHFG. Besides, a novel Self-attentive Refined Attention Module (SAttRAM) is devised and applied in the generator to extract and fuse the features from source image and skeleton image effectively. 2) Extensive experiments were conducted on six different styles of Chinese font generation tasks. Both automatic and human evaluations demonstrate the efficacy of SE-GAN compared to strong baselines. 3) To facilitate future research on BHFG, we also contribute a large-scale font image dataset and will release it soon.

\section{Related Work}
\noindent\textbf{Image-to-Image Translation.} Since the propose of Generative Adversarial Network (GAN) \cite{goodfellow2014generative}, many works have been proposed for image-to-image translation tasks. Pix2pix \cite{pix2pix} is a conditional GAN based on supervised learning with paired data.  Then unsupervised image translation models such as CycleGAN \cite{zhu2017cyclegan} is proposed to use the cycle consistency to improve the training stability. Apart from one-to-one domain image translation, StarGAN \cite{stargan} introduces a domain classifier or a shared multi-domain embedding to achieve many-to-many domain translation with a single model. Furthermore, U-GAT-IT \cite{kim2019ugatit} proves the effectiveness of attention module in image translation task, which can guide the model to focus on more important regions between different domains. However, these works are formulated as pixel-to-pixel translation where the source and target images contain little deformations, which cannot be directly applied to font generation tasks with huge holistic changes.

\noindent\textbf{Chinese Font Generation.} Most previous works formulate character image generation as image translation tasks \cite{zi2zi,emd2018,DG-Font,ecalligan}. Zi2zi \cite{zi2zi} generates stylized Chinese font generation with paired data. However, it requires large-scale font-pair corpus for pre-training and fine-tuning. To overcome the challenge of insufficient data, \cite{chang2018generating} apply cycle consistency to generate fonts from unpaired data, and \cite{emd2018,DG-Font} separate content and style as two irrelevant domains and learn the latent style from font images. Recently, \cite{zeng2021diversity} utilize StarGAN \cite{stargan} equipped with diversity regularizer to achieve multi-style Chinese font generation. Some other works \cite{gaogan,jiang2019scfont,zeng2020strokegan,wen2021stroke} try to integrate more domain knowledge of character into multi-stage model for character generation. Most of these works concentrate on the SPFG, and the images from source and target domain usually share very similar styles or shapes. In this paper, we focus on the BHFG task, which contains large stroke style difference and layout changes. We propose a one-stage model instead of multi-stage model, thus all modules are trained jointly and the generation process is more efficient.

\section{Approach}
\subsection{Overall Framework}
Figure \ref{fig:framework} shows our framework with two encoders: the image encoder $E_i$ and the skeleton encoder $E_s$, which are composed of four residual blocks. The $E_i$ is employed to extract character image features from $X_i$, including content and style information from source domain whereas the skeleton encoder $E_s$ aims to preserve the structure feature from $X_s$. To enhance the feature extraction and fusion of two different features, a novel self-attentive refined attention module (SAttRAM) with two variants, SRAM and CRAM, are stacked sequentially to extract the skeleton enhanced image representations. Then, the generator takes the refined image representations which include both content and style information as input to generate the target style image. Following the adversarial training strategy in GANs, we employ two discriminators. The first $D_i$ is used to discriminate the target image and generated image. The second $D_s$ tries to detect whether the skeleton of generated image is coherent to the skeleton image extracted from the target domain character image. For skeleton extraction, inspired by \cite{fastthining1984}, we adopt an simple but effective skeletonization algorithm to extract the skeleton image by eroding and dilating the binarized character image iteratively.

\subsection{Self-attentive Refined Attention Module}
The previous works \cite{zhou2016cam,kim2019ugatit,lee2019ficklenet} demonstrate the effectiveness of class activation map (CAM) in localizing the important regions of input images in both image generation and classification tasks. Inspired by this, CAM can be used to extract style-discriminative attention heatmaps in the font generation task. To obtain style-discriminative features $M(x)$, the decoded feature maps $F(x) \in \mathbb{R}^{C \times W \times H}$ of source font $x$ are first fed into the classifier, a fully connected (FC) layer with weights $\Omega \in \mathbb{R}^{C}$, for domain classification. Then CAM computes the attention heatmaps by linearly weighted summation of all channels:
\begin{equation}
\begin{aligned}
M(x) =\sum_{c=k}^{C}\Omega_kF(x)_{k}
\end{aligned}
\label{equa:att}
\end{equation}
where $M(x)  \in \mathbb{R}^{W \times H \times C}$ indicates the attention heatmap at spatial location $H, W$, $\Omega_k$ represents the weight for channel $k$ in feature maps, and $F(x)_k \in \mathbb{R}^{W \times H}$ represents the feature map of channel $k$ from the last convolutional layer at spatial location $HW$. 

\begin{figure}[ht]
    \begin{center}
          \includegraphics[width=0.85\linewidth]{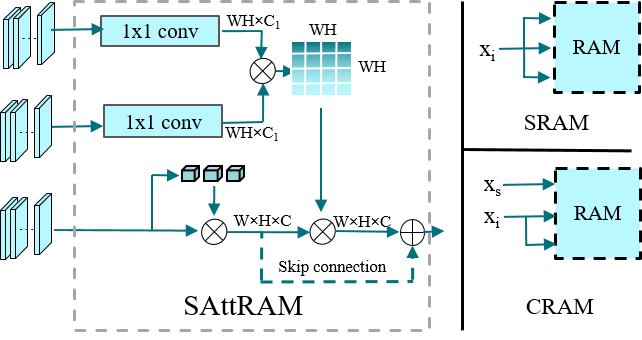}
    \end{center}
    \caption{The proposed SAttRAM. There are two variant modules: SRAM takes one group of input features $X_i$, CRAM takes two groups of input features $X_i$ and $X_s$.}
    \label{fig:ram}
\end{figure}

However, the CAM lacks the spatial attention and usually leads to over-activation issues for feature capturing \cite{chattopadhay2018grad++}. Moreover, it's difficult to integrate multi-modal features with a CAM module. To refine the style-discriminative feature and integrate multi-modal features, we propose a self-attentive refined attention module (SAttRAM) as shown in Fig \ref{fig:ram}. We bring self attention as an efficient module to refine the pixel-wise attention heatmaps by capturing spatial dependency. Hence, the refined feature maps can be defined as follows:
\begin{equation}
\begin{aligned}
\hat{M}(x) =f(F(x),F(x))g(M(x)) + M(x)
\end{aligned}
\label{equa:att}
\end{equation}
\begin{equation}
    \begin{aligned}
    \label{equa:ff}
        f(F(x),F(x)) = \sigma(\theta (F(x))^T)\phi(F(x)))
    \end{aligned}
\end{equation}

Here $f$ is a pairwise embedding function that computes dot-product pixel affinity as self-refined attention weight normalised by softmax function $\sigma$ in an embedding space. The embedding functions $\theta$, $\phi$ are implemented by individual $1 \times 1$ convolution layers, where $\theta (F(x))\in \mathbb{R}^{C1 \times WH}$ and $\phi (F(x))\in \mathbb{R}^{C1 \times WH}$. The function $g$ reshapes the input feature $F(x)$ to $g(F(x))\in \mathbb{R}^{WH}$, all of which are aggregated with this similarity weights given by function $f(F(x),F(x))\in \mathbb{R}^{C \times WH}$. This self-refined attention weight is normalised by softmax function $\sigma$ and the output is $\hat{M}(x)\in \mathbb{R}^{W \times H \times C}$.

To handle the multimodal input features for image generation, we build two variant attention units on top of the refined attention feature maps, namely the self-refined attention module (SRAM) unit and the cross-refined attention module (CRAM) unit. SRAM takes a group of images features $x_i= E_i(x)$ as input to obtain attended features, since image features are basic information in image translation. CRAM catches intra-modal interactions between image features $x_i= E_i(X_i)$ and skeleton features $x_s= E_s(X_s)$, which can further refine the feature maps extracted from character images.
\begin{equation}
    \begin{aligned}
        \label{eqn:ramskeleton}
\hat{M}(x_i, x_s)) =f(F(x_i),F(x_s))g(M(x_i)) + M(x_i)
    \end{aligned}
\end{equation}

In Equation \ref{eqn:ramskeleton}, the character encoding feature is embedded into the residual space by function $g$. Whereas $f$ represents the pixel-level feature aggregation between image feature $x_i$ and skeleton feature $x_s$. 

\begin{table*}[h]
\small
 \caption{Evaluation on font image generation. \textit{Acc} represents content accuracy of character recognition.}
\centering
  \begin{tabular}{l|c|c|c|c|c|c|c|c|c|c|c|c}
    \hline
    Styles &
      \multicolumn{2}{c|}{Style 1} &
      \multicolumn{2}{c|}{Style 2} &
      \multicolumn{2}{c|}{Style 3} &
      \multicolumn{2}{c|}{Style 4} &
      \multicolumn{2}{c|}{Style 5} &
      \multicolumn{2}{c}{Style 6} \\   
    \hline
    Models & Acc  & FID & Acc  & FID &Acc  & FID& Acc  & FID &Acc  & FID&Acc &FID \\
    \hline
    zi2zi \cite{zi2zi}& 0.201&93.32&0.291&84.01&0.443&70.79&0.468&92.52&0.487&81.14&0.362&81.47\\
 CycleGAN \cite{chang2018generating} & 0.348&77.16&0.277&76.05&0.432&76.51&0.542&83.79&0.335&62.64&0.472&79.62\\
    StarGAN \cite{zeng2021diversity}& 0.257&83.68&0.222&93.74&0.437&67.88&0.464&84.66&0.310&92.16&0.321&85.80\\
    DF-Font \cite{DG-Font}& 0.421& 103.82&0.317&96.85& 0.414 & 80.75 &0.515&94.52&\textbf{0.644}&101.71&0.442 & 94.31 \\
    \hline\hline
         SE-GAN & \textbf{0.434}&\textbf{62.58}&\textbf{0.486}&\textbf{60.43}&\textbf{0.513}&\textbf{50.40}&\textbf{0.628}&\textbf{61.22}&0.616&\textbf{54.69}&\textbf{0.532}&\textbf{73.33} \\
         SE-GAN w/o $E_s$ & 0.406&76.51&0.369&80.15&0.475&66.87&0.589&78.88&0.542&70.63&0.435&83.04 \\
         SE-GAN w/o $D_s$ & 0.427&72.14&0.388&74.86&0.491&57.64&0.621&70.52&0.539&66.19&0.507&80.61 \\ 
    \hline\hline
      Human&0.465&--&0.521&--&0.604&--&0.638&--&0.812&--&0.568&-- \\ \hline
  \end{tabular}
  \label{table:accuracy}
\end{table*}

\subsection{Discriminators}
\label{sect:discriminator}
We design two discriminators for learning target font style. Following the traditional setting, the first discriminator $D_i$ is employed to calculate how similar the generated image is to the font image of target domain, and the encoder for $D_i$ also exploits the refined attention feature maps as mentioned in previous section. Under the assumption that the generated character image should have similar skeleton to its target font image, we also design an extra skeleton discriminator $D_s$. Considering that the skeletonization is a non-differential function, we first apply a pre-trained skeleton generator (skeletonGAN) to generate the corresponding skeleton from a given character image. Then the skeleton discriminator $D_s$ is applied to distinguish the skeleton of generated image from that of ground truth image. For skeletonGAN, we train a simple CycleGAN model \cite{zhu2017cyclegan} without SAttRAM module, as skeletonization task appears to be much easier than the character generation task. 
With the help of skeleton discriminator $D_s$, the model is encouraged to preserve the content and structure of character during training, meanwhile, $D_s$ is also served as a regularizer for SE-GAN to prevent model from overfitting. 

\subsection{Loss Design}
The loss function of SE-GAN contains content loss, adversarial loss, cycle-consistency loss and classification loss. The content loss consists $L_{con}$ of two parts: the pixel loss $L_{pix}$ which forces the generated image $G_F(X_i,X_s)$ to be similar to the target image $Y_t$, and the skeleton consistency loss $L_{sc}$ which ensures the skeletons consistency between $SG(G_F(X_i,X_s))$ and $SG(Y_i)$.
\begin{equation}
\begin{aligned}
L_{pix}(G_F) = \mathbb{E}_{X}[||G_F(X_i,X_s) -Y_t ||_1] 
\end{aligned}
\end{equation}
\begin{equation}
\begin{aligned}
L_{sc}(G_F,SG) = \mathbb{E}_{X}[||SG(G_F(X_i,X_s))-SG(Y_t)||_1] 
\end{aligned}
\end{equation}
where the $SG$ represents the pre-trained skeletonGAN.

The cycle-consistency loss $L_{cycle}$ is identical to that used in \cite{zhu2017cyclegan}, which guarantees that the cycle transformation is able to bring the image back to the original state. 
In order to distinguish that the image $X_i$ belongs to source or target style $y_{cls}$ and promote the style transformation in the refined attention module, we use an auxiliary loss $L_{cls}$ following \cite{kim2019ugatit}.

Besides, the adversarial loss $L_{adv}$ combines discriminative loss $L_{D_i}$ and skeleton-level discriminative loss $L_{D_s}$,
and these two losses aim to catch different properties of the desired generated image $X_i$.
\begin{equation}
\begin{aligned}
L_{D_i}(G_F,D_i) = 
& \mathbb{E}_{Y}[logD_{i}(G_F(X_i,X_s))] \\
& +  \mathbb{E}_{X}[log(1-D_{i}(G_F(X_i,X_s)))] 
\end{aligned}
\end{equation}
\begin{equation}
\begin{aligned}
L_{D_s}(G_F,D_{s}) = 
& \mathbb{E}_{Y}[logD_{s}(Y_s)] \\
& +  \mathbb{E}_{X}[log(1-D_{s}(SG(G_F(X_i,X_s))))] 
\end{aligned}
\end{equation}
where $D_i$, $D_s$ are pixel-level discriminator and skeleton discriminator respectively.




Finally, we jointly train the generators, discriminators, and classifiers by using the full objective as follows:
\begin{equation}
\begin{aligned}
\underset{G_F,G_B,RAM_p}{min}\underset{D_{1},D_{2}}{max} \lambda_1{L}_{adv}&+\lambda_2{L}_{cycle}+\lambda_3L_{cls}+\lambda_4L_{con}
\end{aligned}
\label{equt:loss}
\end{equation}
where $\lambda_i$ controls the weights of different losses. Here, we omit the corresponding backward loss functions for simplicity because they are also defined in the same way.

\section{Experiments}
\begin{table}[h]
\small
\caption{Statistics of six styles in our dataset.}
\centering
\begin{tabular}{l|ccccccc}
  \hline
  Style &S1 & S2 & S3 & S4 & S5 &S6 \\ \hline
   Number &1885&3008&3958&1419 &2356 &3175  \\
  \hline
\end{tabular}
\label{table:user}
\end{table}
\subsection{Experimental Setup}
As the deficiency of public brush handwriting font generation dataset, we collect a large-scale image dataset for experiments with six different styles.
The statistics of each style is presented in Table 1. The total number of images is 15,799, and the size of each subset ranges from 1,419 to 3,958. During experiments, we split each subset into train/dev/test set by ratio of 8:1:1. We choose the standard print font Liukai\footnote{\url{https://www.foundertype.com/index.php}} as the source domain, and take each of the six styles as the target domain respectively. 
We adopt content accuracy \cite{calligraphy_cycleGAN} and Fréchet Inception Distance (FID) \cite{FID2017} as evaluation metrics. For baselines, we compare our model with four representative font generation models, including zi2zi \cite{zi2zi}, CycleGAN \cite{chang2018generating}, StarGAN \cite{zeng2021diversity}, and DGFont \cite{DG-Font}. 

\begin{figure*}[htb]
	\centering 
	\includegraphics[width=0.8\linewidth]{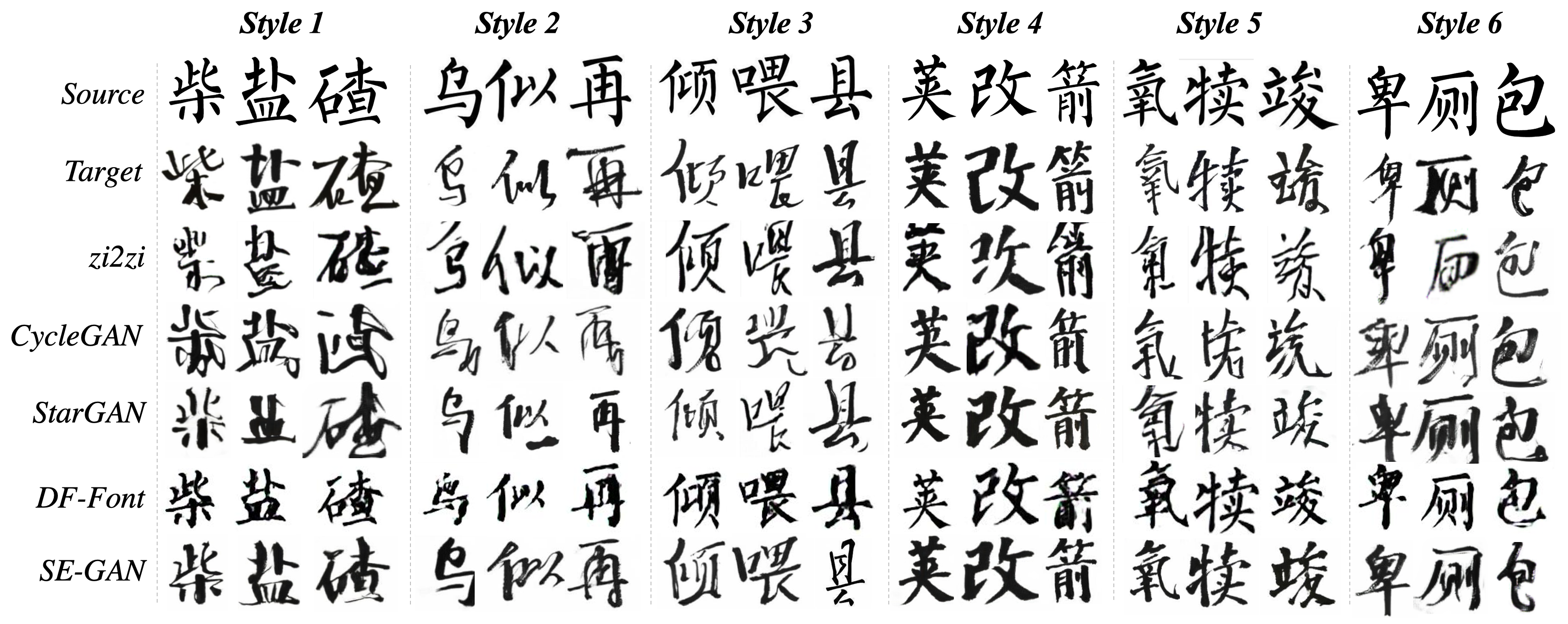}
	\caption{Comparison of generated font images from different baselines and SE-GAN.
	}
	\label{fig:chars}   
\end{figure*}

\subsection{Experimental Results}

\noindent\textbf{Quantitative analysis.} We calculate both the Top 1 content accuracy (Acc) and FID score for all baselines and our proposed model on six chirography styles. Table \ref{table:accuracy} illustrates that our proposed model SE-GAN achieves the best accuracy and the lowest FID score for nearly all six styles. Compared with supervised methods like zi2zi, our model can still generate high-quality images. Compared with unsupervised models CycleGAN and StarGAN, which have unstable performances among different styles, SE-GAN still has very competitive performance and significant improvements. For DF-Font, we notice that it is inclined to generate similar font images to the source images, however, the distinctive styles are not well learnt. Although the content accuracy of DF-Font is high, the FID score is the worst.

\noindent\textbf{Qualitative analysis.} We show some generated examples of all models in Fig. \ref{fig:chars} for case study. Generally, the font images from SE-GAN are much easier to recognize and the chirography styles are more consistent with the original styles mentioned. For other baselines, there exists obvious flaws in the generated font images. For zi2zi, there are missing strokes for Style 2 and Style 6, and the structures are tied to each other for Style 1 and Style 5. For CycleGAN, the structures are wrong and the characters are hard to recognize for Style 1 and 3. For StarGAN, there are some extra and erroneous strokes in Style 2 and 3, which damage the overall appearance of characters. All above examples demonstrate that the skeleton information can facilitate the font generation.

\begin{figure}[t]
	\centering 
	\includegraphics[width=1\linewidth]{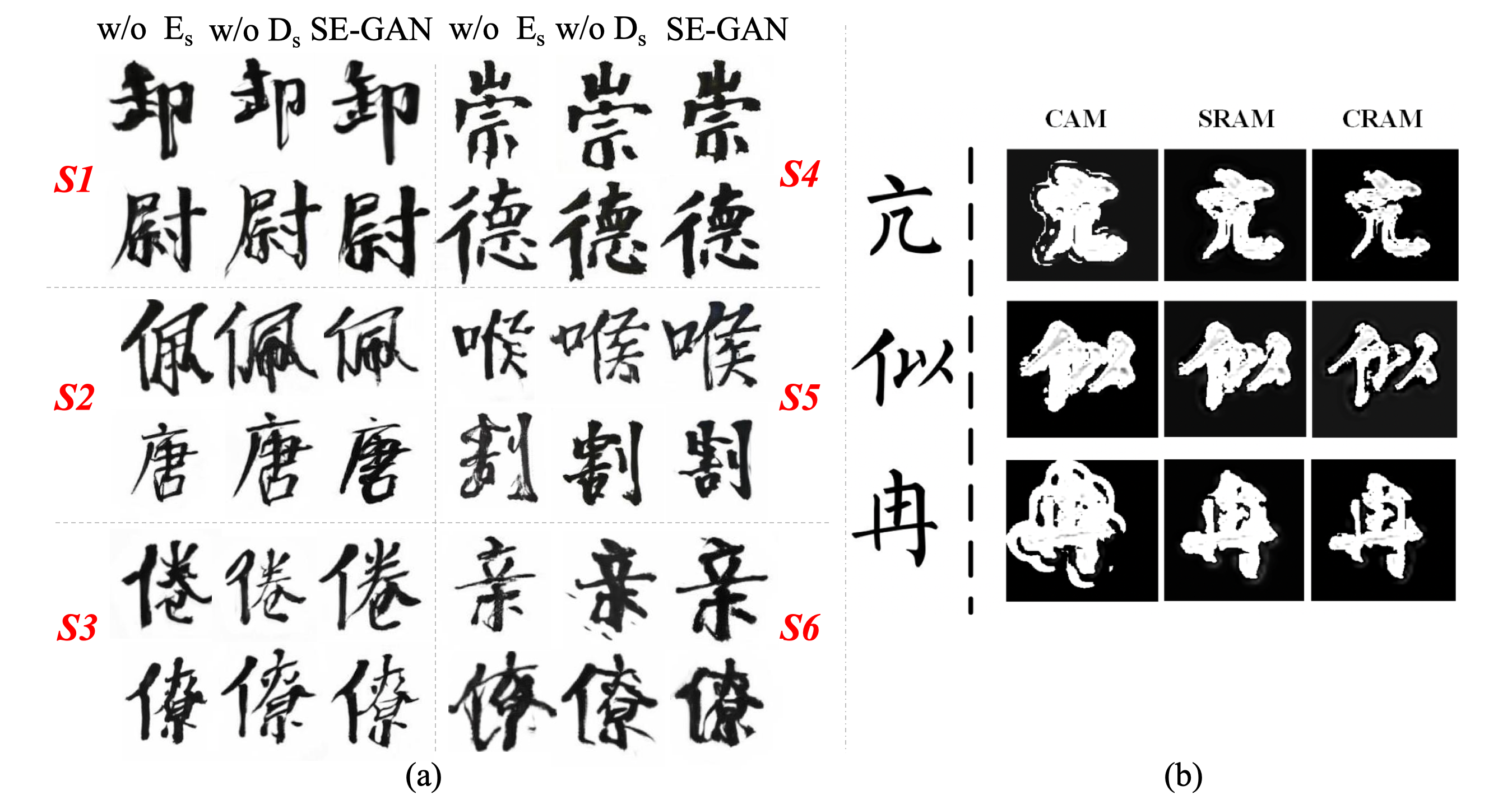}
	\caption{(a) Examples of generated font images with ablated models; (b) Visualization of different attention modules.}  

	\label{fig:abla}   
\end{figure}
\noindent\textbf{Ablation Study.} We conduct ablation experiments by removing the skeleton input $E_s$ from generator and removing the skeleton discriminator $D_s$ separately. As shown in Table \ref{table:accuracy}, after removing $E_s$, the accuracy drops evidently compared with SE-GAN, which indicates the skeleton information is helpful for learning the structure of characters. However, it still outperforms CycleGAN for most of the styles, proving the effectiveness of SAttRAM. Besides, removing $D_s$ also degrades the model performance over all styles, indicating the necessity of skeleton discriminator. Fig. \ref{fig:abla} (a) illustrates the generated images of different ablated models. Due to the lack of skeleton information, some strokes in the generated characters are either missing or over exaggerated. When $D_s$ is removed, some components are mixed together, hurting the readability of the generated images. Fig. \ref{fig:abla} (b) compares the heated attention maps of CAM, SRAM and CRAM. It's observed that, compared with CAM, SRAM and CRAM have fewer over-activations and more complete activation coverage. Besides, the font shape learnt by CRAM is more accurate than SRAM and closer to the ground-truth font image, which verified the contribution of skeleton.


\noindent\textbf{User Study.} We conduct two kinds of user study (we skip evaluating DF-Font considering its bad performance in Table \ref{table:accuracy}). The first is to evaluate preference score for the generated font images of different models. The second is to pick out the more visually pleasing font image by mixing the generated images with human-written font images. Totally sixty students majored in fine arts with more than 3-year experience of calligraphy writing were invited to finish the human evaluation. Table \ref{table:user} reveals our model obtains the highest user preference score and winning rate from the human experts.
\begin{table}[h]
\small
\caption{The user study results of different models.}
\centering
\begin{tabular}{l|ccccc}
  \hline
  Models & zi2zi & CycleGAN & StarGAN & SE-GAN \\ \hline
  Pref score & 2.3 & 3.1 &2.9 &\textbf{4.2}  \\
  Win rate & 14\% & 18\% & 35\% & \textbf{57\%} \\
  \hline
\end{tabular}
\label{table:user}
\end{table}

\section{Conclusion and future work}
In this paper, we propose SE-GAN, a novel end-to-end framework for brush handwriting font generation. As the task involves holistic structure changes and complex stroke transfer, we propose to integrate the skeleton information for character image generation. We design two encoders to extract the character and skeleton features respectively. To efficiently fuse the information from both sides, a novel self-attentive attention module is devised in the generator. Besides, we also employ a skeleton discriminator to ensure the content consistency between generated and target images. The experiments demonstrate the advantages of our proposed model over several strong baseline methods. In the future, we will explore the pre-trained image translation models to facilitate this task.

\bibliographystyle{IEEEbib}
\bibliography{sample-base}

\end{document}